\pdfoutput=1

\documentclass[11pt]{article}

\usepackage{EACL2023}

\usepackage{times}
\usepackage{latexsym}

\usepackage[T1]{fontenc}

\usepackage[utf8]{inputenc}

\usepackage{microtype}

\usepackage{inconsolata}

\usepackage{amsmath,amsfonts,bm}

\def\eqref#1{equation~\ref{#1}}

\def\1{\bm{1}}

\def\vc{{\bm{c}}}

\def\vt{{\bm{t}}}

\def\vv{{\bm{v}}}

\def\vx{{\bm{x}}}
\def\vy{{\bm{y}}}
\def\vz{{\bm{z}}}

\def\mI{{\bm{I}}}

\DeclareMathAlphabet{\mathsfit}{\encodingdefault}{\sfdefault}{m}{sl}
\SetMathAlphabet{\mathsfit}{bold}{\encodingdefault}{\sfdefault}{bx}{n}

\def\gF{{\mathcal{F}}}

\def\gL{{\mathcal{L}}}

\usepackage{hyperref}
\usepackage{url}
\usepackage{multirow,booktabs,makecell}
\usepackage{booktabs}
\usepackage{graphicx}
\usepackage{adjustbox}
\usepackage{color}
\usepackage{comment}
\usepackage[labelformat=simple]{subcaption}

\usepackage[T1]{fontenc}
\usepackage{enumitem}
\usepackage{mathabx}
\usepackage{bm}
\usepackage{soul}
\usepackage{amssymb}
\usepackage{pifont}

\usepackage{colortbl}
\newcommand{\cmark}{\ding{51}}%
\newcommand{\xmark}{\ding{55}}%

\newcommand{\methodname}[1]{iNLG}
\newcommand{\iv}[1]{I\&V}

\title{Visualize Before You Write: \\ Imagination-Guided Open-Ended Text Generation}

\author{
  Wanrong Zhu\textsuperscript{\P},
  An Yan\textsuperscript{\dag},
  Yujie Lu\textsuperscript{\P},
  Wenda Xu\textsuperscript{\P}, 
  \\
  \textbf{Xin Eric Wang\textsuperscript{\S}}, 
  \textbf{Miguel Eckstein\textsuperscript{\P}},
  \textbf{William Yang Wang\textsuperscript{\P}}
  \\
  \textsuperscript{\P}UC Santa Barbara, 
  \textsuperscript{\dag}UC San Diego, 
  \textsuperscript{\S}UC Santa Cruz
  \\
  {\small \{wanrongzhu,yujielu,wendaxu,william\}@cs.ucsb.edu, ayan@ucsd.edu} \\
  {\small xwang366@ucsc.edu, miguel.eckstein@psych.ucsb.edu}  \\
}

\begin{document}
\maketitle


\begin{abstract}
Recent advances in text-to-image synthesis make it possible to visualize machine imaginations for a given context. 
On the other hand, when generating text, human writers are gifted at creative visualization, which enhances their writings by forming imaginations as blueprints before putting down the stories in words. 
Inspired by such a cognitive process, we ask the natural question of whether we can endow machines with the same ability to utilize visual information and construct a general picture of the context to guide text generation. 
In this work, we propose \methodname~ that uses machine-generated images to guide language models (LM) in open-ended text generation. 
The experiments and analyses demonstrate the effectiveness of \methodname~ on open-ended text generation tasks, including text completion, story generation, and concept-to-text generation in both few-shot and full-data scenarios.
Both automatic metrics and human evaluations verify that the text snippets generated by our \methodname~ are coherent and informative while displaying minor degeneration.\footnote{Our code \& data: \url{https://github.com/VegB/iNLG}.}

\end{abstract}

\section{Introduction}

One great resource human writers cherish is the ability of imagination, with which they render mental images about an actual or vicarious experience and link knowledge that would later make the writing more concrete, sensible, and intriguing.
Cognitive studies show that visual imagery improves comprehension during language processing~\citep{gambrell1986mental,joffe2007comprehension,sadoski2013imagery}, and that mental imagery facilitates humans' written language expression at young ages~\citep{gambrell2002imagery}.

\begin{figure}[t]
    \centering
    \includegraphics[width=\linewidth]{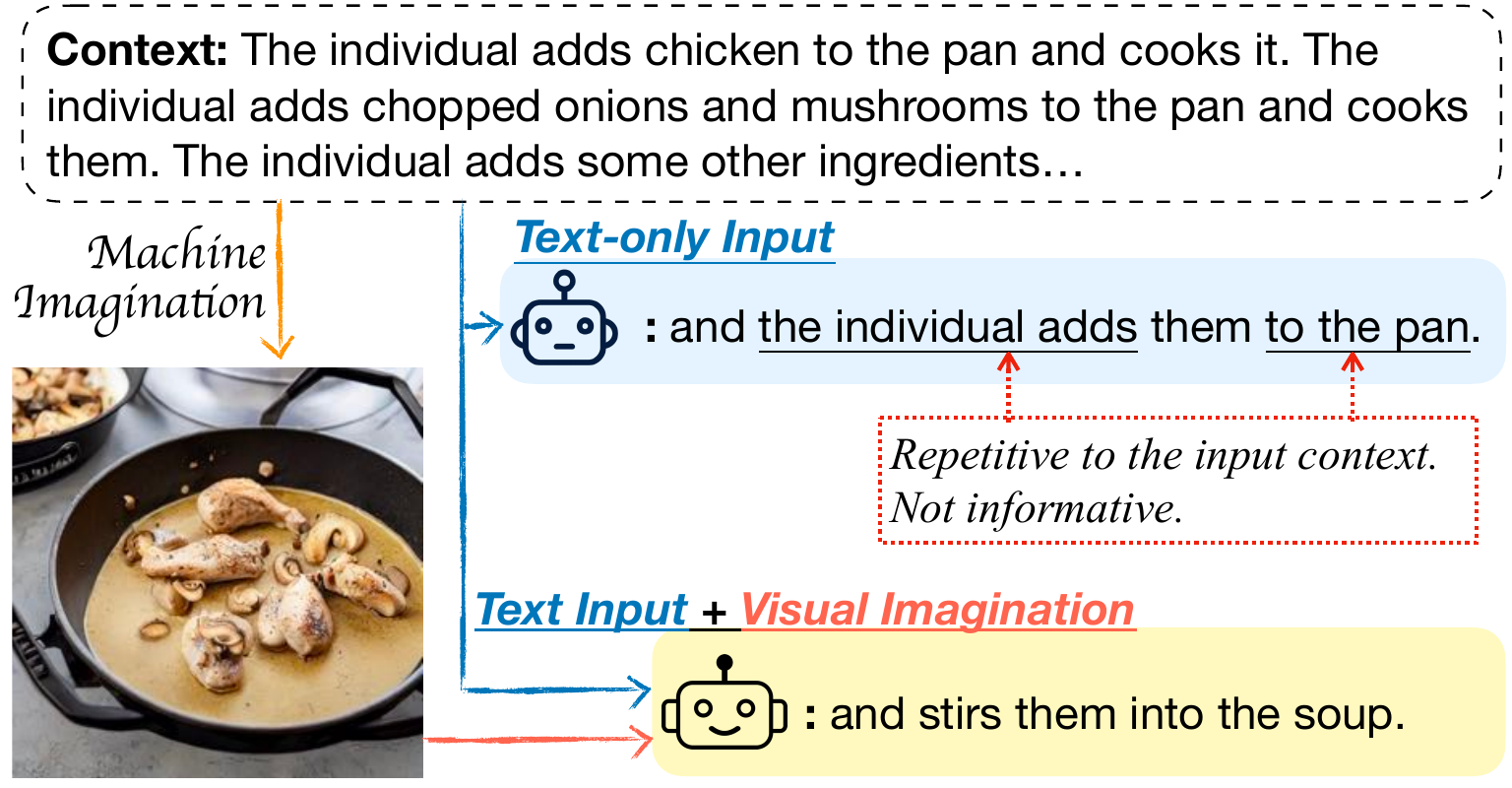}
\vspace{-20px}
\caption{
When performing open-ended text generation, the LMs prompted with text-only input may generate repetitive or unilluminating contents, which is also known as degeneration. Hereby, we propose to use machine-generated images as additional visual supervision to guide the LMs in generating more informative and coherent text with the given context.
}
\vspace{-10px}
\label{fig:intro_eg}
\end{figure}

When it comes to the study of Artificial Intelligence (AI), one classic challenge for AI systems is to generate informative and coherent text snippets.
Open-ended text generation is such a task that provides an input context, and asks the model to generate a piece of text that is consistent with the context.
This is the cornerstone of a wide range of downstream tasks such as text completion~\citep{Guan2019StoryEG,Radford2019LanguageMA}, story generation~\citep{fan-etal-2018-hierarchical,goldfarb-tarrant-etal-2020-content,swanson-etal-2021-story,Su2022ACF}, and dialogue systems~\citep{schatzmann-etal-2007-agenda,wen-etal-2015-semantically,wen-etal-2017-network,wei-etal-2018-task,wu-etal-2021-spoken}, and has received much attention throughout the years. 
Inspired by human writers' common practice of creative visualization, we ask the following question: Can we endow machines with the same ability to construct a general picture of the context and use it as a blueprint to guide text generation?

Recent advances in text-to-image generation make it possible to visualize machine imaginations for a given context~\citep{Ramesh2021ZeroShotTG,Rombach2022HighResolutionIS,Crowson2022VQGANCLIPOD,Wang2022UnifyingAT,Saharia2022PhotorealisticTD}. Moreover, this line of work shows great potential in utilizing textual information to guide image synthesis. It comes naturally that one may attempt to complete the loop by using visual supervision to guide text generation.

In this work, we propose using machine-generated images to guide the language model (LM) in open-ended text generation. 
More specifically, we visualize machine imagination for the input context by rendering images with StableDiffusion~\citep{Rombach2022HighResolutionIS}, a state-of-the-art text-to-image generator.
The machine imagination acts as additional visual supervision to guide LMs in generating informative and coherent text in two ways.
Firstly, the machine-generated images are introduced as the input to the LM in the form of the visual prefix.
Secondly, we designed a contrastive training objective that enforces the generated text to be semantically similar to the visual supervision.

We conduct experiments on three open-ended text generation tasks, namely text completion, story generation, and concept-to-text generation.
Extensive experiments in the few-shot settings show better or competitive performance to state-of-the-art baselines on both automatic metrics and human evaluation.
Experiments with full-data settings show that introducing machine-generated visual supervision with our \methodname~ yields consistent improvements on various LM models including GPT-2~\citep{Radford2019LanguageMA}, BART~\citep{lewis-etal-2020-bart}, and T5~\citep{2020t5}.

Our main contributions are as follows:
\begin{itemize}[noitemsep,topsep=0pt,parsep=0pt,partopsep=0pt]
    \item We introduce a novel paradigm that leverages machine-generated images to guide open-ended text generation. This endows the machines with the ability of creative visualization that human writers often demonstrate.
    \item We distill the vision information from the pre-trained multimodal models and further construct visual prefixes to guide language models performing text generation with teacher forcing and contrastive objectives. 
    \item Extensive experiments show the effectiveness of \methodname~ as a model-agnostic framework in open-ended text generation tasks, including text completion, story generation, and concept-to-text in both few-shot and full-data settings.
\end{itemize}

\section{Related Work}

\paragraph{Open-ended Conditional Text Generation} is the task of generating a coherent portion of the text based on the given context. Recent advances in pre-trained models have pushed frontier in the open-ended conditional text generation, such as text completion\cite{see-etal-2019-massively, ippolito-etal-2020-toward}, story generation \cite{https://doi.org/10.48550/arxiv.2001.05139, fan-etal-2018-hierarchical, Yao2019PlanAndWriteTB} and concept-to-text generation \cite{ZhouLSL021, Liu2021KGBARTKG}. Despite the success of large language models, text degeneration and semantic coverage still remain as two core technical challenges in few-shot open-ended text generation. To improve the text coverage, StoryEndGen \cite{Guan2019StoryEG} leverages the knowledge graph to encode context sequentially. \citet{fan-etal-2018-hierarchical} and \citet{Yao2019PlanAndWriteTB} plan the content (premise or keywords) first and then encourage the generation based on planned content. To mitigate the text degeneration, SimCTG \cite{Su2022ACF} uses a contrastive training strategy to encourage the model to learn isotropic token embeddings. 
Similar to our approach, \citet{wang2022contextualized} generates a scene graph for each concept and combines them with text for the model input. 
Previous work has proposed to add visual information to LM by retrieving images from the Internet or large-scale image sets~\citep{yang2020visual,pmlr-v139-cho21a,Su2022LanguageMC}. However, the retrieved images may fail to fully incorporate the context, which will misguide the LM from yielding contextually consistent predictions.\footnote{Figure~\ref{fig:retrieved_vs_generated} shows examples where the image retrieved from the search engine is irrelevant with the input context.}
Unlike prior work, our approach leverages images generated conditioning on the context to assist the text generation process.

\paragraph{Visually-aided NLP}
Recent work show the power of visual guidance in natural language processing, spanning from the language representation learning~\citep{Lu2019ViLBERTPT,Li2019VisualBERTAS,Sun2019VideoBERTAJ,Luo2020UniViLMAU,Chen2020UNITERUI,Li2020UnicoderVLAU,Tan2020VokenizationIL,Yujie2022iACE}, the downstream tasks~\citep{Grubinger2006TheIT,Elliott2016Multi30KME, Xie2019VisualEA, Christie2016ResolvingLA, Shi2019VisuallyGN, Yujie2022iACE} and evaluation~\citep{Zhu2021ImaginEAI}.
They either leverage visual information from an external vision-and-language corpus or obtain such visual knowledge from the large pre-trained model.
In this line of work, imagination achieves promising performance in various NLP domains~\citep{long2021imt, Zhu2021ImaginEAI, wang2022contextualized, Yujie2022iACE}.
Previous imagination-based work in NLP either study non-generation problems~\citep{Zhu2021ImaginEAI, Yujie2022iACE} or utilize non-visual information~\citep{long2021imt, wang2022contextualized}. 
Our work explores the potential of generating visual imagination to improve open-ended text generation tasks.


\begin{figure*}[t]
    \centering
    \includegraphics[width=\linewidth]{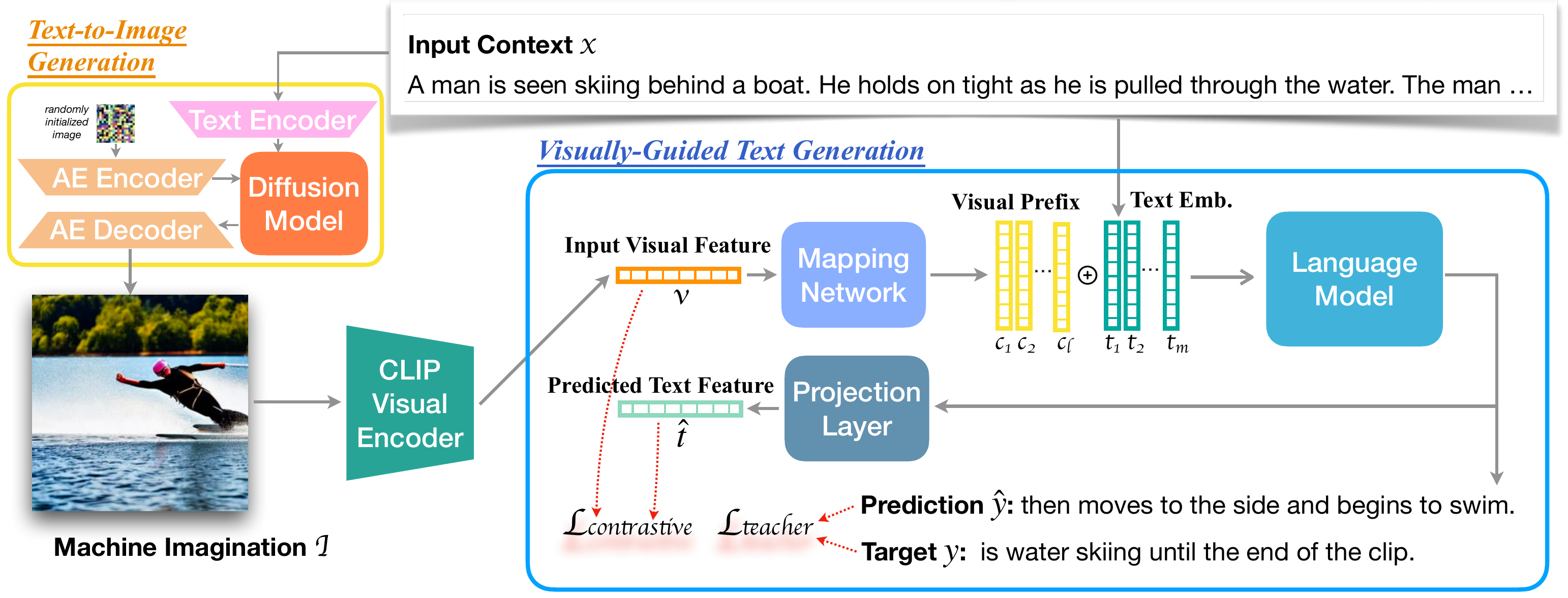}
\caption{
An overview of our \methodname~.
Given an input context $\vx$, we first visualize the context with the text-to-image generation model.
Then we use the machine-generated image $\mI$ as the additional visual supervision to guide the language model in open-ended text generation.
The visual feature is provided as a source of input to the LM in the form of the visual prefix. 
Aside from the teacher forcing objective $\gL_{\text{teacher}}$, we also enforce the LM to generate text that is semantically similar to the machine imagination with a contrastive training objective $\gL_{\text{contrastive}}$.
}
\label{fig:model_overview}
\end{figure*}

\section{Method}

\subsection{Overview}

Open-ended text generation is a task that provides an input context, and asks the model to generate a piece of text that is consistent with the context. 

This work mainly focused on introducing machine-rendered images to assist LM in performing open-ended text generation. More specifically, given the context $\vx^i$, we first use a text-to-image generator to illustrate an image $\mI^i$ that depicts the input context. The LM is prompted with image $\mI^i$ as the visual prefix along with the text context $\vx^i$, and will incorporate the multimodal input to generate the output text $\hat{\vy^i}$.

Figure~\ref{fig:model_overview} provides an overview of our \methodname~ framework, which mainly involves two modules.
The first module is a text-to-image generator that takes in the input context and illustrates a descriptive image, which we also refer to as the machine imagination.
The second module is a visually-guided language model that utilizes the machine imagination as a source of input and also a supervision that encourages the LM to generate text that is semantically similar to the visual information.

\subsection{Text-to-Image Rendering}


In this work, we propose to use images generated conditioning on the context by the machines as additional visual information to the LM. 
The text-to-image generation backbone is StableDiffusion~\citep{Rombach2022HighResolutionIS}, which mainly consists of a text encoder, a diffusion model, and an autoencoder.
The text encoder is from the frozen CLIP ViT-L/14~\citep{Radford2021LearningTV} and encodes the input text to textual embeddings.
The diffusion model uses UNet~\citep{10.1007/978-3-319-24574-4_28} to provide noise estimation. The UNet is modified so as to attend to the input textual embeddings.
The encoder of the pretrained autoencoder encodes images into the lower-resolution latent maps $\vz_T$. At each step $t$, the diffusion model provides the noise estimation $\epsilon$ and modifies $\vz_t$ correspondingly.
The decoder of the pretrained autoencoder takes the final noise-free latent map $\vz$ and generates the image prediction. StableDiffusion is trained with LAION-5B~\citep{schuhmann2022laionb}.



\subsection{Visually Guided Text Generation}
\paragraph{Visual Prefix Construction}
One can encode the visual information with the pre-trained visual models. However, such visual embedding may lie in a representation space different from the LM due to the discrepancy between models. 
One way of introducing features extracted by another network to the current model is through feature mapping~\citep{mokady2021clipcap}. 
With a dataset of image-text pairs $(\mI', \vx')$, we can pre-train a mapping network $\gF$ for a given LM in an image captioning formulation.
More specifically, we encode $\mI'$ with the visual encoder $\text{Enc}_{\text{visual}}$ and receive its visual features $\vv'$. Then we apply the mapping network $\gF$ over $\vv'$, and receive a sequence of $l$ visual prefixes:
\begin{equation}
    c'_1, c'_2, \dots, c'_l = \gF(\vv') = \gF(\text{Enc}_{\text{visual}}(\mI'))
\end{equation}
We provide the list of visual prefix as input to the LM with the corresponding text $\vx'$ as the target output. Such a pre-training process enables $\gF$ to project visual features into the visual prefix that lies within the same embedding distributions as the LM. The mapping network is agnostic of the downstream task, and only depends on the visual source and the LM.

After generating a descriptive image $\mI^i$ for the input context $\vx^i$, we use CLIP to encode $\mI^i$ and receive its visual features $\vv^i$. We apply the pre-trained mapping network $\gF$ over $\vv^i$, and receive the visual prefix $\vc^i$ of length $l$:
\begin{equation}
   \vc^i = \{c_1^i, c_2^i, \dots, c_l^i\} = \gF(\text{CLIP}(\mI^i))
\end{equation}

\paragraph{Visually-guided Language Modeling}
We use the visual information to guide text generation in two ways, reflected in the following two training objectives.
Firstly, we directly introduce the machine-generated visual information as input to the LM. We concatenate the visual prefix $\vc^i$ and the text embeddings $\vt^i$ for the input context $\vx^i$ with $m$ tokens. LM input can be denoted as $[\vc^i;\vt^i] = \{c_1^i,  \dots, c_l^i, t_1^i,\dots, t_m^i\}$.
With $\vy^i = \{y_1^i, y_2^i, \dots, y_n^i\}$ denoting the target output of $n$ tokens, and $\theta$ denoting the trainable parameters, we can list out the teacher forcing training objective as follows:
\begin{equation}
    \gL_{\text{teacher}} = - \sum_{j=1}^{n} \log p_{\theta}(y^i_j|\vc^i;\vt^i;\vy^i_{<j})
\end{equation}

In addition, we design a contrastive objective to enforce the generated text to be semantically similar to the input visual supervision with the InfoNCE loss~\citep{Oord2018RepresentationLW, yan2021weakly}: 
\begin{equation}
    \gL_{\text{contrastive}} = - \log \frac{\exp(\mathrm{sim}(\vv^i,\hat{\vt^i})/\tau)}{\sum_{j\neq i}\exp(\mathrm{sim}(\vv^i,\hat{\vt^j})/\tau)}
\end{equation}
in which $\hat{\vt}$ is the projected representation of the decoder's last layer's output, and can be viewed as the sentence-level representation of the generated text.
Here $\mathrm{sim}(\cdot,\cdot)$ first normalizes the two vectors, then compute their cosine similarity, and $\tau$ is the temperature.  

\subsection{Training \& Inference}

We first pre-train the mapping network on the pre-training dataset with the teacher-forcing objective. Such pre-training is agnostic of the downstream task, and only depends on the type of base LM. 

When applying our \methodname~ on downstream tasks, we train the base LM with the teacher forcing objective for the first $N_{\text{no\_contra}}$ epochs.
Then, we introduce the contrastive objective and tune the base LM together with the mapping network and projection layer by minimizing the following loss $\gL$. Here $ep$ denotes the epoch and $\lambda$ is the factor:
\begin{equation}
    \gL = 
    \left\{
        \begin{aligned}
        &\gL_{\text{teacher}},  &ep < N_{\text{no\_contra}}, \\
        &\gL_{\text{teacher}} + \lambda \gL_{\text{contrastive}}, &ep > N_{\text{no\_contra}}, \\
        \end{aligned}
    \right.
\end{equation}

During inference, we provide the context and machine-generated image to the LM. We use beam search during decoding with a beam width of 10.
\begin{table*}[!htp]\centering
\begin{adjustbox}{width=\linewidth,center}
\begin{tabular}{cll}\toprule
\textbf{Task} &\textbf{Input Context} &\textbf{Target Output} \\

\cmidrule[\heavyrulewidth]{1-3}

Text Completion &
\makecell[l]{
Different people are interviewed on camera while several others
\\are shown raking up the leaves. A man is seen sitting in his car
\\and another puts his gloves on. The camera }&
\makecell[l]{
pans over the raked up leaves while several others discuss their
\\hard work.} \\
\cmidrule{1-3}

Story Generation &
Live Show. Tim was in his school's play. &
\makecell[l]{
He was nervous about their first show. He almost dropped out.
\\The show went smoothly. Tim was excited for his second show.
} \\
\cmidrule{1-3}

Concept-to-Text &
grow, flower, pavement &
Wild flower growing through crack in the tiled pavement. \\
\bottomrule

\end{tabular}
\end{adjustbox}
\caption{Exemplars of the input context and corresponding target output for three open-ended text generation task covered in this study, namely story generation, text completion, and concept-to-text generation.}
\label{tab:data_example}
\end{table*}

\section{Experimental Setup}

\subsection{Tasks, Datasets, and Baselines}
We apply our \methodname~ on three open-ended text generation setups: sentence completion, story generation, and concept-to-text generation. Table~\ref{tab:data_example} shows examples for each task.

\paragraph{Sentence Completion} is a task of finishing the sentence in a commonsense inference scenario.
We conduct experiments on the ActivityNet~\citep{Heilbron2015ActivityNetAL} subset\footnote{14740/982/2261 samples for train/validation/test.} of HellaSwag~\citep{zellers-etal-2019-hellaswag}, which is a benchmark for commonsense natural language inference that ask the model to predict the most likely follow-up among several choices given a specific context. 
We compare with StoryEndGen~\citep{Guan2019StoryEG} which encodes the given context incrementally and attends to the one-hop knowledge graph retrieved from ConceptNet for the context tokens. 
We implement our ~\methodname~ on top of the GPT-2~\citep{Radford2019LanguageMA}, which by nature, can generate the follow-up for an arbitrary input in a zero-shot manner.

\paragraph{Story Generation} requires the model to compose a story based on the given title or context. We conduct experiments on the widely used story generation benchmark ROCStories~\citep{mostafazadeh-etal-2016-corpus}. Each data item consists of a story title and a human-written five-sentence everyday life story that incorporates commonsense related to the title.\footnote{We use the split provided by ~\citet{Su2022LanguageMC}, which is based on the ROCStories Winter 2017 release and contains 49666/1500/1500 items for the train/validation/test sets.}
We provide the story title and the story's first sentence as the input context, and ask the LM to predict the following four sentences.
We consider the following methods as baselines:
Action-Plan~\citep{fan-etal-2018-hierarchical} first predicts the premise of a story with the convolutional LM~\citep{Dauphin2017LanguageMW}, then use the fusion mechanism~\citep{DBLP:conf/interspeech/SriramJSC18} to encourage a convolutional seq2seq model~\citep{Gehring2017ConvolutionalST} to generate the story from the premise.
Plan-and-Write~\citep{Yao2019PlanAndWriteTB} first plans a storyline that consists of keywords, then generate the story conditioned on the storyline. Its model structure is built upon GRU~\citep{Cho2014OnTP}.
SimCTG~\citep{Su2022ACF} proposes a contrastive training objective that encourages the LM to learn discriminative and isotropic token representations, and is implemented on GPT-2~\citep{Radford2019LanguageMA}.

\paragraph{Concept-to-Text} is a relatively more constrained conditional text generation task involving commonsense reasoning. This task provides a set of concepts as input, and requires the model to generate a piece of text that incorporates the concepts and describes an everyday scenario. We conduct experiments on the CommonGen~\citep{lin-etal-2020-commongen} benchmark.\footnote{We use the in-house split provided by \citet{wang2022contextualized}, which contains 65323/2066/4018 samples for train/validation/test.}
We compare against the following models:
KG-BART~\citep{Liu2021KGBARTKG} encompasses the relations of concepts with the knowledge graph and augments the BART~\citep{lewis-etal-2020-bart} encoder and decoder with graph representations.
ModelAdapt~\citep{ma-etal-2021-exploring} is built upon BART and removes the positional embedding in the encoder.
Imagine-and-Verbalize (\iv~)~\citep{wang2022contextualized} predicts a scene graph for each set of concepts, and uses it as an additional input to the LM. In contrast to \iv~, we directly visualize the concepts and use the machine-generated images as the auxiliary information to assist the concept-to-text generation.

\begin{table*}[t]\centering
\begin{adjustbox}{width=\linewidth,center}
\begin{tabular}{crlrrrrrrr}\toprule
\textbf{Task} &\textbf{*} &\textbf{Setting} &\textbf{rep-2~$\downarrow$} &\textbf{rep-3~$\downarrow$} &\textbf{rep-4~$\downarrow$} &\textbf{diversity~$\uparrow$} &\textbf{distinct-2~$\uparrow$} &\textbf{MAUVE$\uparrow$} &\textbf{BERTScore$\uparrow$} \\
\cmidrule[\heavyrulewidth]{1-10}
\multirow{5}{*}{\makecell[c]{Sentence\\Completion}} 
    &   \texttt{0} &Human &0.45 &0.05 &0.01 &99.50 &77.32 &- &- \\
    &   \texttt{1} &GPT2 \textit{no finetune}~\citep{Radford2019LanguageMA} &6.71 &6.87 &10.13 &78.07 &74.83 &44.19 &22.57 \\
\cmidrule{2-10}
    &   \texttt{2} &StoryEndGen~\citep{Guan2019StoryEG} &39.53 &35.11 &39.30 &34.12 &44.57 &0.45 &-47.29 \\
    &   \texttt{3} &GPT2 \textit{text-only finetune} &4.20 &4.03 &5.53 &86.85 &75.14 &49.45 &24.13 \\
    \rowcolor{gray!10}&   \texttt{4} &GPT2 $+$\methodname~ &\textbf{2.43} &\textbf{2.61} &\textbf{3.57} &\textbf{91.63} &\textbf{75.92} &\textbf{60.30} &\textbf{24.25} \\
\cmidrule[\heavyrulewidth]{1-10}
\multirow{7}{*}{\makecell[c]{Story\\Generation}} 
    &   \texttt{5} &Human &1.76 &0.38 &0.15 &97.71 &56.34 &- &- \\
    &   \texttt{6} &GPT2 \textit{no finetune} &37.65 &22.76 &21.92 &45.67 &43.42 &0.43 &-7.77 \\
\cmidrule{2-10}
    &   \texttt{7} &Action-Plan~\citep{fan-etal-2018-hierarchical} &52.05 &35.58 &28.11 &26.97 &21.43 &0.41 &-18.32 \\
    &   \texttt{8} &Plan-and-Write~\citep{Yao2019PlanAndWriteTB} &45.22 &32.86 &23.34 &30.71 &20.83 &0.41 &-37.35 \\
    &   \texttt{9} &SimCTG~\citep{Su2022ACF} &28.72 &24.02 &20.61 &43.00 &42.06 &0.43 &18.01 \\
    &   \texttt{10} &GPT2 \textit{text-only finetune} &25.41 &18.51 &14.41 &52.10 &46.60 &9.10 &21.23 \\
    \rowcolor{gray!10}&   \texttt{11} &GPT2 $+$\methodname~ &\textbf{10.73} &\textbf{5.64} &\textbf{3.42} &\textbf{81.36} &\textbf{51.91} &\textbf{35.94} &\textbf{23.03} \\

\bottomrule
\end{tabular}
\end{adjustbox}
\caption{
Generation quality scores for few-shot text completion on the ActivityNet and few-shot story generation on ROCStories.
``Human'' shows the human performance and ``GPT2 \textit{no finetune}'' denotes the vanilla GPT2 model without tuning.
All the other listed models are trained with 1\% of the training data. 
``$+$\methodname~'' denotes introducing machine-generated images on top of the base LM.
}
\label{tab:act_roc_auto_scores}
\end{table*}

\subsection{Evaluation}
\paragraph{Automatic}
For sentence completion and story generation, we follow previous work and evaluate the quality of the generated text from the aspect of model degeneration level (rep-$n$, diversity, distinct-$n$), text distribution divergence (MAUVE), and semantic similarity (BERTScore):
(1) rep-$n$ = 1.0 - $\frac{|\text{unique }n\text{-grams}|}{|\text{total }n\text{-grams}|}$ measures sequence level repetition by computing the portion of duplicate $n$-grams~\citep{DBLP:conf/iclr/WelleckKRDCW20}. 
(2) diversity = $\prod_{n\text{=2}}^{4}(1-\text{rep-}n)$ measures the diversity of $n$-grams~\citep{Su2022LanguageMC}.
(3) distinct-$n$ = $\frac{|\text{unique }n\text{-grams}|}{|\text{length of text}|}$ measures the portion of distinct $n$-grams in the text~\citep{Li2016ADO}. 
(4) MAUVE measures the learned distributions divergence between the generated text and human-written text~\citep{DBLP:conf/nips/PillutlaSZTWCH21},\footnote{We report MAUVE with gpt2-large as the base model.} a low MAUVE indicates a great difference between the distributions of generated text and human text.
(5) BERTScore assesses contextual text similarity between two pieces of texts by computing the cosine similarities between their tokens' embeddings~\citep{bert-score},\footnote{We report BERTScore with roberta-large as base model.} a low BERTScore means the generated text is contextually different from the ground-truth.

For concept-to-text, following prior work, we report the metrics scores on BLEU~\citep{Papineni2002BleuAM}, METEOR~\citep{Banerjee2005METEORAA}, CIDEr~\citep{Vedantam2015CIDErCI}, SPICE~\citep{Anderson2016SPICESP}, and BERTScore~\citep{bert-score}.

\paragraph{Human}
We also set up a human evaluation as a complementary evaluation beyond the automatic metrics. We select 100 samples from the test set for sentence completion and story generation and perform the head-to-head comparison between the text snippets generated by our \methodname~ and the baseline models. We invite human annotators to compare the text quality from the following three independent aspects:
(1) \textit{Coherence}: Which snippet is more semantically consistent with the context, and follows the logic of the context more naturally.
(2) \textit{Fluency}: Which snippet is more fluent in English.
(3) \textit{Informativeness}: Which snippet contains more interesting content, and describes the scenes that are more likely to happen in real life.
Three human judges rate each comparison.

\subsection{Implementation Details}
We use StableDiffusion-\texttt{v1-1}~\citep{Rombach2022HighResolutionIS} to render a 512x512 image from the context, and use CLIP ViT/B-32 to extract features offline. 
The mapping network is an 8-layer Transformer, and the visual prefix length is 20. 
For the sentence completion and story generation tasks, 
the mapping network is pre-trained on the MSCOCO~\citep{Lin2014MicrosoftCC} dataset. 
For the concept-to-text task, 
the mapping network is pre-trained on VIST~\citep{huang-etal-2016-visual}.\footnote{CommonGen is built upon image and video captioning datasets including MSCOCO. To avoid data leakage, we choose to pre-train the mapping network on VIST, which is not revealed to CommonGen.} 
We pre-train the mapping network for 5 epochs with a batch size of 128.
Results are reported on three repeat runs.
Detailed hyperparameters are listed in the Appendix.


\begin{table*}[!htp]\centering
\begin{adjustbox}{width=\linewidth,center}
\begin{tabular}{clrrrrrrrrrr}\toprule
\multirow{2}{*}{\textbf{Task}} &\multirow{2}{*}{\textbf{Models}} &\multicolumn{3}{c}{\textbf{Coherence}} &\multicolumn{3}{c}{\textbf{Fluency}} &\multicolumn{3}{c}{\textbf{Informativeness}} \\
\cmidrule(lr){3-5}\cmidrule(lr){6-8}\cmidrule(lr){9-11}
& &Win(\%) &Tie(\%) &Lose(\%) &Win(\%) &Tie(\%) &Lose(\%) &Win(\%) &Tie(\%) &Lose(\%) \\
\cmidrule[\heavyrulewidth]{1-11}
\multirow{3}{*}{Sentence Completion} &Ours vs. StoryEndGen &\textbf{51.67} &20.33 &28.00 &\textbf{44.67} &19.33 &36.00 &\textbf{41.33} &18.33 &40.33 \\
&Ours vs. GPT2 \textit{no finetune} &\textbf{51.00} &22.67 &26.33 &\textbf{45.00} &22.33 &32.67 &\textbf{41.00} &21.00 &38.00 \\
&Ours vs. GPT2 \textit{text-only finetune} &\textbf{58.00} &24.33 &17.67 &\textbf{43.33} &18.67 &38.00 &\textbf{42.33} &21.67 &36.00 \\\cmidrule{1-11}
\multirow{5}{*}{Story Generation} &Ours vs. Action-Plan &\textbf{51.00} &24.67 &24.33 &\textbf{54.67} &16.33 &29.00 &\textbf{52.00} &15.00 &33.00 \\
&Ours vs. Plan-and-Write &\textbf{45.33} &25.67 &29.00 &\textbf{53.00} &16.67 &30.33 &\textbf{54.67} &17.00 &28.33 \\
&Ours vs. SimCTG &\textbf{42.00} &27.67 &30.33 &\textbf{40.33} &25.67 &34.00 &\textbf{43.33} &18.33 &38.33 \\
&Ours vs. GPT2 \textit{no finetune} &\textbf{43.33} &24.33 &32.33 &\textbf{43.67} &20.33 &36.00 &\textbf{44.67} &19.00 &36.33 \\
&Ours vs. GPT2 \textit{text-only finetune} &\textbf{39.33} &26.67 &34.00 &\textbf{38.67} &26.67 &34.67 &\textbf{44.33} &22.67 &33.00 \\
\bottomrule
\end{tabular}
\end{adjustbox}
\vspace{-10px}
\caption{Human evaluation results for the sentence completion task and the story generation task. The scores indicate the percentage of win, tie or lose when comparing our \methodname~ with the baseline models. 
}

\label{tab:human_eval_act_roc}

\end{table*}

\section{Result and Analysis}

\subsection{Few-Shot Learning Results}

Open-ended text generation is a broad topic with flexible and inexhaustible setups, many of which have low resources. Collecting annotations is often extremely expensive and time-consuming. Therefore, we first report few-shot results to check if our \methodname~ can rapidly adapt to new task setups with a few examples, which is more practical in real-life.

More specifically, we report few-shot open-ended text generation results with 1\% of the training data.
For sentence completion and story generation tasks, the base LM is GPT2-base~\citep{Radford2019LanguageMA}. For concept-to-text, we test it with BART-base~\citep{lewis-etal-2020-bart} as the base LM.

\paragraph{Sentence Completion} 
As shown in Table~\ref{tab:act_roc_auto_scores}, StoryEndGen (\texttt{\#2}) suffers from degeneration with the highest rep-$n$ and the lowest diversity. Training with only 1\% of the training data improves GPT2's performance on all metrics (\texttt{\#3} vs. \texttt{\#1}).
Under the same few-shot setting, adding additional machine-generated images with our \methodname~ (\texttt{\#4}) further alleviate model degeneration. The improvement on MAUVE also indicates that introducing visual input can aid GPT2 in generating text that is more similar to the human-written ones.

\paragraph{Story Generation} 
As shown in Table~\ref{tab:act_roc_auto_scores}, for the story generation task that requires the LM to compose longer text, we see the vanilla GPT2 without tuning suffering from more severe degeneration compared to rendering a sentence ending (\texttt{\#6} vs. \texttt{\#1}).
The high rep-$n$ scores indicate that the two non-Transformer-based baselines Action-Plan (\texttt{\#7}) and Plan-and-Write (\texttt{\#8}) stammer with repetitive tokens, which greatly differs from the human-written text (leads to low MAUVE) and does not have concrete meanings (leads to low BERTScore). 
The models based on GPT-2 (\texttt{\#9}-\texttt{\#10})
yield more complete sentences with concrete meanings (BERTScore gets higher). However, they keep repeating the same sentence, which is still quite different from human language (MAUVE remains low). 
Applying \methodname~ to GPT-2 leads to minor degeneration and has the best performance on all metrics (\texttt{\#11}).
Examples of generated text snippets can be found in Figure~\ref{fig:showcase_mainpaper_rocstories} and in Appendix.

\begin{table}[t]
\begin{adjustbox}{width=\linewidth,center}
\begin{tabular}{clrrrrr}\toprule
\textbf{*} &\textbf{Setting} &\textbf{B-4} &\textbf{M.} &\textbf{CIDEr} &\textbf{SPICE} &\textbf{BertS.} \\\cmidrule{1-7}
\texttt{1} &BART-base \textit{text-only finetune} &20.72 &25.47 &114.49 &24.58 &59.76 \\
\texttt{2} &$+$KG~\citep{Liu2021KGBARTKG}  &15.26 &24.44 &98.53 &23.13 &52.76 \\
\texttt{3} &$+$Adapt~\citep{ma-etal-2021-exploring} &23.11 &25.96 &123.44 &25.14 &61.53 \\
\texttt{4} &$+$\iv~~\citep{wang2022contextualized} &24.50 &25.89 &119.61 &25.59 &57.29 \\
\rowcolor{gray!10}\texttt{5} &$+$\methodname~ &\textbf{25.07} &\textbf{26.48} &\textbf{127.93} &\textbf{26.32} &\textbf{63.37} \\
\bottomrule
\end{tabular}
\end{adjustbox}
\caption{Automatic metrics scores for few-shot concept-to-text generation on CommonGen with 1\% of the training data. 
All listed models are implemented on BART-base.
``$+$KG'' adds knowledge graph, ``$+$Adapt'' applies model adaption, ``$+$\iv~'' adds scene graph, and ``$+$\methodname~'' introduces machine-generated images as input. 
B-4: BLEU-4; M.: METEOR; BertS.: BERTScore.
}
\label{tab:commongen_auto_scores}
\end{table}

\paragraph{Concept-to-Text}
Table~\ref{tab:commongen_auto_scores} shows that knowledge graph information may not be fully exploited under the few-shot setting (\texttt{\#2}), while removing the information of relative positions between input concepts helps the LM write better sentences (\texttt{\#3}).
Introducing machine-generated images can improve the base LM's performance on concept-to-text generation (\texttt{\#5} vs. \texttt{\#1}).
While both \iv~ and our \methodname~ involve machine ``imagination'', we provide such information in different forms (scene graphs vs. images).
Comparing \texttt{\#4} and \texttt{\#5}, our \methodname~ outperforms \iv~ with BART-base as the base LM.
This suggests that the additional information introduced by \iv~ and \methodname~ is complementary. 

\paragraph{Human Evaluation} Table~\ref{tab:human_eval_act_roc} lists out human evaluation results on text completion and story generation. Our \methodname~ outperforms the compared baselines on all three criteria in the model-level head-to-head comparisons.
This further verifies the effectiveness of our \methodname~ in generating fluent and informative text snippets that better align with the given context.

\subsection{Model-Agnostic Improvement}

We further report open-ended text generation results with various base LM when trained with the full set of data.
For concept-to-text, we experiment with BART-base/large~\citep{lewis-etal-2020-bart} and T5-base/large~\citep{2020t5}.
For sentence completion and story generation, we record results on GPT2-base/large~\citep{Radford2019LanguageMA}.
As shown in Table~\ref{tab:full_set_results}, introducing machine-generated visual supervision with our \methodname~ leads to model-agnostic improvements over text-only finetuning. This holds true for all the listed base LM with different architectures and verifies that our \methodname~ is a model-agnostic framework. 


\begin{table}[t]
\begin{adjustbox}{width=\linewidth,center}
\begin{tabular}{llrrrrrr}\toprule
\textbf{Base LM} &\textbf{Setting} &\multicolumn{5}{c}{\textbf{Metrics}} \\
\cmidrule[\heavyrulewidth]{1-7}
\rowcolor{gray!8}\multicolumn{2}{c}{\textbf{\textit{Concept-to-Text}}} &\textbf{B-4$\uparrow$} &\textbf{MET.$\uparrow$} &\textbf{CIDEr$\uparrow$} &\textbf{SPICE$\uparrow$} &\textbf{BertS.$\uparrow$} \\
\cmidrule{1-7}
\multirow{2}{*}{BART-base} &text-only &30.32 &31.35 &158.92 &31.22 &68.50 \\
&$+$\methodname~ &\textbf{30.60} &\textbf{31.44} &\textbf{160.63} &\textbf{31.42} &\textbf{69.02} \\
\cmidrule(lr){1-7}
\multirow{2}{*}{BART-large} &text-only &32.38 &33.06 &169.69 &33.01 &70.33 \\
&$+$\methodname~ &\textbf{32.76} &\textbf{33.17} &\textbf{171.47} &\textbf{33.35} &\textbf{70.79} \\
\cmidrule(lr){1-7}
\multirow{2}{*}{T5-base} &text-only &30.39 &30.87 &163.67 &32.77 &70.03\\
&$+$\methodname~ &\textbf{31.09} &\textbf{31.18} &\textbf{165.52} &\textbf{32.81} &\textbf{70.35}\\
\cmidrule(lr){1-7}
\multirow{2}{*}{T5-large} &text-only &34.13 &32.91 &175.67 &34.30 &72.44 \\
&$+$\methodname~ & \textbf{34.50} &\textbf{33.87} &\textbf{177.65} &\textbf{35.48} &\textbf{72.70} \\
\cmidrule[\heavyrulewidth]{1-7}
\rowcolor{gray!8}\multicolumn{2}{c}{\textbf{\textit{Sentence Completion}}} &\textbf{rep-4$\downarrow$} &\textbf{div.$\uparrow$} &\textbf{dist-2$\uparrow$} &\textbf{MAUVE$\uparrow$} &\textbf{BertS.$\uparrow$} \\
\cmidrule{1-7}
\multirow{2}{*}{GPT2-base} &text-only &4.20 &87.46 &72.87 &61.42 &29.84 \\
&$+$\methodname~ &\textbf{3.95} &\textbf{89.33} &\textbf{74.09} &\textbf{64.01} &\textbf{30.10} \\
\cmidrule(lr){1-7}
\multirow{2}{*}{GPT2-large} &text-only &\textbf{1.77} &\textbf{96.54} &76.74 &87.81 &31.66  \\
&$+$\methodname~ &2.05 &95.90 &\textbf{76.80} &\textbf{89.11} &\textbf{32.15} \\
\cmidrule[\heavyrulewidth]{1-7}
\rowcolor{gray!8}\multicolumn{2}{c}{\textbf{\textit{Story Generation}}} &\textbf{rep-4$\downarrow$} &\textbf{div.$\uparrow$} &\textbf{dist-2$\uparrow$} &\textbf{MAUVE$\uparrow$} &\textbf{BertS.$\uparrow$} \\
\cmidrule{1-7}
\multirow{2}{*}{GPT2-base} &text-only & 7.83 &68.42 &49.53 &33.13 &28.81 \\
&$+$\methodname~ & \textbf{6.80} &\textbf{71.17} &\textbf{49.92} &\textbf{38.86} &\textbf{29.13} \\
\cmidrule(lr){1-7}
\multirow{2}{*}{GPT2-large} &text-only &1.02 &91.91 &54.17 &82.81 &31.86 \\
&$+$\methodname~ &\textbf{0.85} &\textbf{92.51} &\textbf{54.54} &\textbf{87.83} &\textbf{32.03} \\
\bottomrule
\end{tabular}
\end{adjustbox}

\caption{
Automatic metric scores when trained with the full set of data with ablations of the base LM.
Introducing our \methodname~ leads to model-agnostic improvements across the board.
B-4: BLEU-4; MET.: METEOR; BertS.: BERTScore; div.: diversity; dist-2: distinct-2.
}
\label{tab:full_set_results}
\end{table}

\subsection{Performance Analysis}

\paragraph{Source of Image}
We first perform an ablation study to understand how the source of visual information affects our \methodname~ framework.
We compare retrieved/generated images from four sources: (1) the first returned result by Yahoo Image Search;\footnote{\href{https://images.search.yahoo.com/}{https://images.search.yahoo.com/}} (2) images rendered by VQGAN+CLIP~\citep{Crowson2022VQGANCLIPOD};\footnote{\href{https://github.com/nerdyrodent/VQGAN-CLIP}{https://github.com/nerdyrodent/VQGAN-CLIP}} (3) images rendered by OFA~\citep{Wang2022UnifyingAT},\footnote{\href{https://github.com/OFA-Sys/OFA}{https://github.com/OFA-Sys/OFA}} and (4) images rendered by StableDiffusion~\citep{Rombach2022HighResolutionIS}, with which we report the main results. 

\begin{figure}[t]
    \centering
    \includegraphics[width=\linewidth]{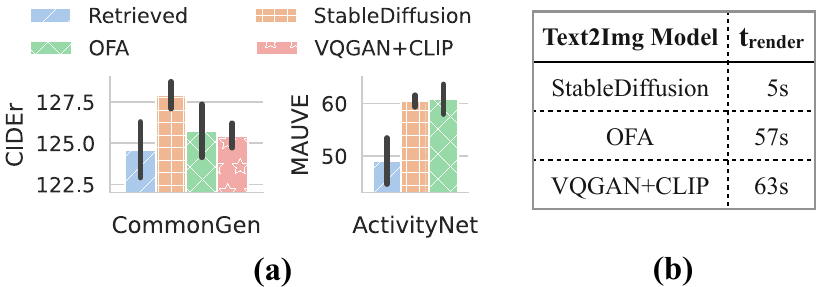}
\vspace{-20px}
\caption{
(a) \methodname~'s performance on CommonGen and ActivityNet with visual supervisions retrieved from the web or generated by machines. Scores are reported with error bars.
(b) Average time to render an image on TITAN RTX
with each text-to-image generator.
}
\label{fig:ablation_image_source}
\vspace{-1ex}
\end{figure}

As shown in Figure~\ref{fig:ablation_image_source}(a), the images generated by machines act as a more effective supervision than the retrieved images. This validates our motivation of introducing machine-generated images over retrieved ones to guide LM in performing text generation. 
Among the three text-to-image generators, VQGAN+CLIP is slightly inferior to the other two, while StableDiffusion and OFA have mixed performance. Images generated by StableDiffusion rank first on CommonGen, while images rendered with OFA score slightly higher on ActivityNet. Figure~\ref{fig:ablation_image_source}(b) reports the average image rendering time, where StableDiffusion is 10$\times$ faster when rendering images than the other two.

\begin{figure}[t]
    \centering
    \includegraphics[width=\linewidth]{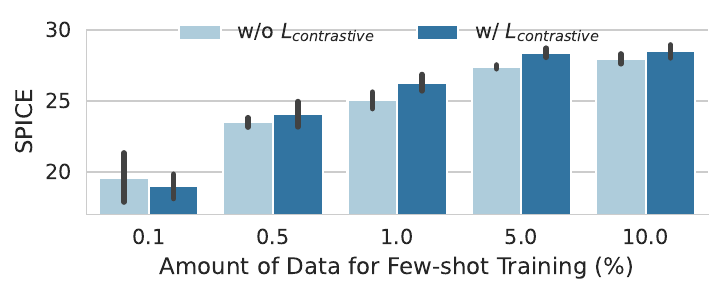}
\vspace{-25px}
\caption{
Performance of applying our \methodname~ on BART-base for few-shot concept-to-text with ablated training objective $\gL_{\text{contrastive}}$ on various few-shot settings. Scores are reported with error bars.
}

\label{fig:ablation_contrastive}
\end{figure}

\paragraph{Contrastive Training}
We examine the effect of the contrastive training objective on CommonGen, and the results are presented in Figure~\ref{fig:ablation_contrastive}. 
We notice that introducing $\gL_{\text{contrastive}}$ improves \methodname~'s performance on 4 out of 5 listed few-shot setups, which suggests that our contrastive training objective generally can assist the LM in composing open-ended text snippets.
One exception is in the extreme few-shot setting with only 0.1\% of training data, where the amount of data is insufficient to let the LM form a decent representation.
In this case, enforcing the sentence representation to be similar to the visual supervision with $\gL_{\text{contrastive}}$ might misguide the LM.

\begin{figure}[t]
    \centering
    \includegraphics[width=\linewidth]{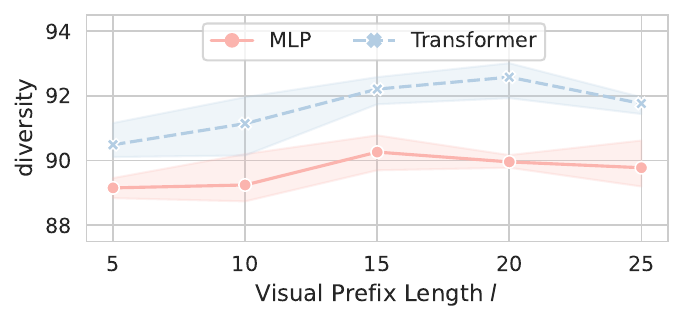}
\vspace{-25px}
\caption{
Performance of our \methodname~ on few-shot sentence completion with various visual prefix lengths and with MLP and Transformer as mapping network. Scores are reported with error bands.
}
\label{fig:ablation_mapping_network}
\end{figure}

\paragraph{Mapping Network \& Visual Prefix}
We discuss the effects of different types of mapping networks and various visual prefix lengths.
Aside from the 8-layer Transformer we used in the main experiments, we also tried a simple Multi-Layer Perceptron (MLP) with two fully-connected layers.
As shown in Figure~\ref{fig:ablation_mapping_network}, the Transformer-based mapping network outperforms MLP on all listed $l$. MLP has the best performance when visual prefix length $l=15$, while the Transformer-based mapping network scores highest when $l=20$.

\paragraph{Model Weight Tuning}
\begin{table}[t]
\begin{adjustbox}{width=\linewidth,center}
\begin{tabular}{cccrrr}\toprule
\textbf{Tune LM} &\textbf{Pretrain Map.} &\textbf{Tune Map.} &\textbf{diversity~$\uparrow$} &\textbf{MAUVE~$\uparrow$} \\\cmidrule{1-5}
\xmark &\xmark &\xmark &15.52 &0.47 \\
\xmark &\xmark &\cmark &78.20 &33.79 \\
\xmark &\cmark &\xmark &27.06 &1.83 \\
\xmark &\cmark &\cmark &76.36 &25.15 \\
\cmark &\xmark &\xmark &87.45 &48.06 \\
\cmark &\xmark &\cmark &88.68 &51.81 \\
\cmark &\cmark &\xmark &89.05 &55.61 \\
\cmark &\cmark &\cmark &92.68 &60.62 \\
\bottomrule
\end{tabular}
\end{adjustbox}
\caption{Performance of our \methodname~ on few-shot sentence completion with ablated settings on whether to tune the LM, pretrain the mapping network (Pretrain Map.) and tune the mapping network (Tune Map.).}
\label{tab:ablation_lm_mapping_tuning}
\end{table}
Table~\ref{tab:ablation_lm_mapping_tuning} compares the influence of pre-training/tuning the weights of different modules of our \methodname~.
Generally speaking, tuning the weights during training outperforms freezing the weights, which applies to both the base LM and the mapping network. In addition, considering our few-show setup, pre-training the mapping network also helps our \methodname~ gain better performances.
The best combination is applying the pre-trained mapping network, and tuning it together with the base LM on the few-shot downstream task.

\begin{figure}[t]
    \centering
    \includegraphics[width=\linewidth]{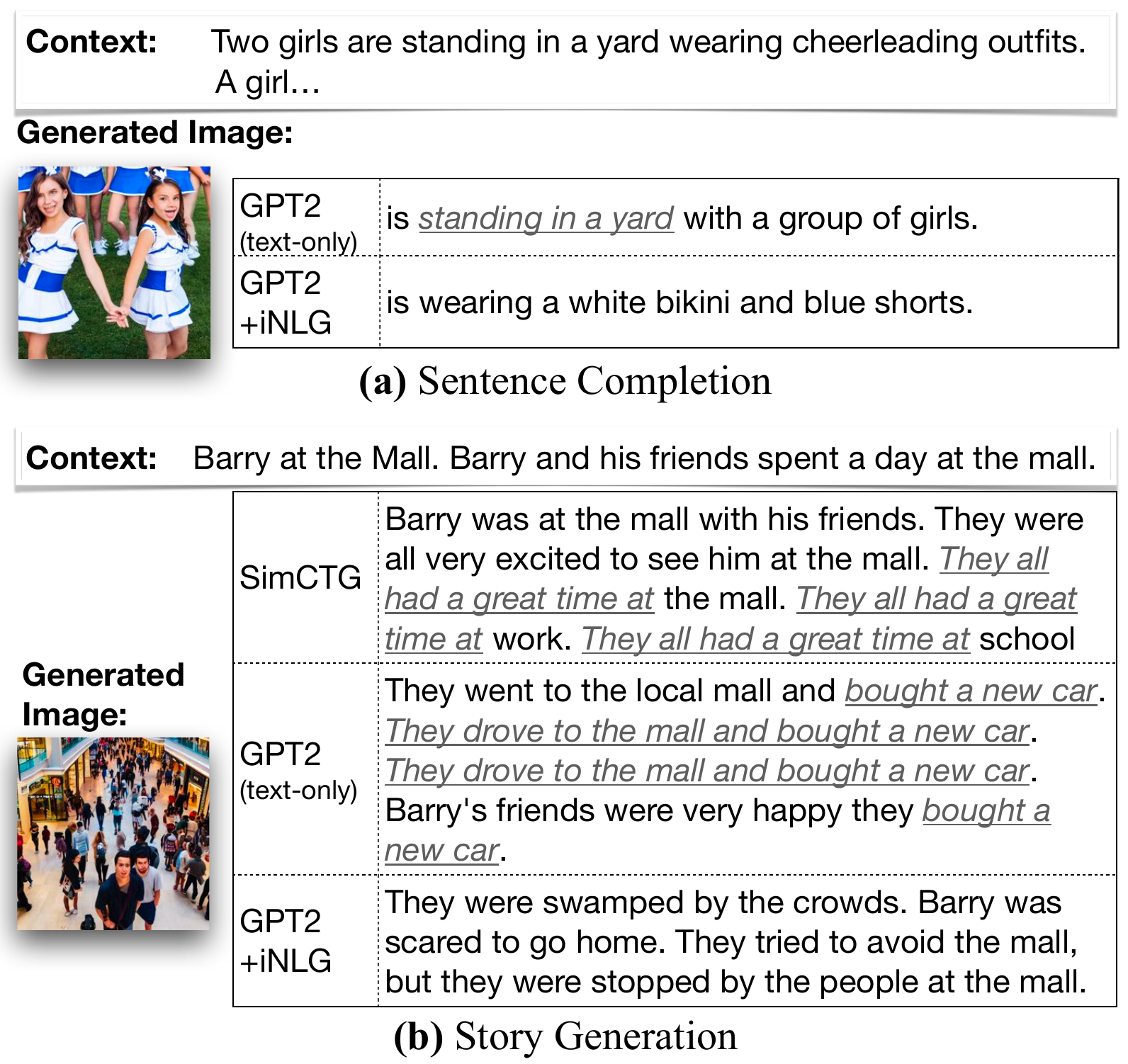}
\vspace{-20px}
\caption{
Sentence ending and stories generated by GPT2-based methods tuned with 1\% of the training data. 
\textit{\underline{Repetitive contents}} are underlined.
The sentence ending and story written by our \methodname~ is coherent with the context, related to the machine-generated image, and has minor degeneration.
More demonstrative examples can be found in the Appendix.
}
\vspace{-10px}
\label{fig:showcase_mainpaper_rocstories}
\end{figure}

\paragraph{Showcase}
Figure~\ref{fig:showcase_mainpaper_rocstories} provides two showcases on few-shot sentence completion and story generation to compare our \methodname~ with the GPT2-based baselines. SimCTG and GPT2 tuned with text-only corpus rendering repeated segments, either copying from the input context, or simply repeating themselves. In comparison, our \methodname~ has minor degeneration and writes coherent sentence endings or stories with more creative details in both tasks.



\section{Conclusion}
In this work, we propose \methodname~, a framework that introduces machine-generated images to guide open-ended text generation. This endows the machines with the ability of creative visualization that human writers often demonstrate.
We distill the vision information from the pre-trained multimodal models and further construct visual prefixes to guide language models to perform text generation with the teacher forcing and the contrastive objective. 
Extensive experiments show the effectiveness of \methodname~ in open-ended text generation tasks, including text completion, story generation, and concept-to-text generation in few-shot settings.


\section*{Limitations}


This work mainly focuses on open-ended text generation, where the search space for the target output is infinite, and the language model would benefit from additional visual imagination distilled from large text-to-image generation models to produce coherent and meaningful content. However, we should note here that despite the commendable performance of text-to-image generation models, there are certain terms and concepts that are inherently challenging to visualize, such as numerical values and abstract philosophical terms. This problem itself is an interesting open research question for all tasks involving text-and-vision.
In our current approach, the images are generated offline. In future work, one may explore the integration of text-to-image and image-to-text modules in an end-to-end manner, which may be more suitable for longer text generation that is not covered in this work.
Text-to-image generation models currently have a length limit on the input text prompt, which may impede their ability to visualize long text inputs in a single image. Furthermore, as previously discussed, text-to-image models may also encounter difficulties in generating images of complex scenes or situations that are challenging to depict through a single image. Future research could explore the use of multiple images or supplementary videos as visual input in order to provide a more comprehensive representation of the scene or situation in question. The \methodname~ framework can be easily extended to take video representation by taking longer visual prefixes or iteratively applying visual prefixes at each step.




\section*{Ethics Statement}

In this work, we use pre-trained multimodal models to visualize machine imagination. The machine-generated images may contain uncontrolled bias if any inductive bias exists from the pre-training data. Even though we do not witness such an issue in our study, this may be a potential factor that affects the quality of the generated text.
We do not anticipate any major ethical concerns given that all the datasets and models used in this study have already been released to the public. We reproduce baselines with the released code repository. 
For human evaluation, our study is approved for IRB exempt. The estimated hourly wage paid to MTurk annotators is \$10.

\subsection*{Acknowledgement}
The research was sponsored by the U.S. Army Research Office and was accomplished under Contract Number W911NF-19-D-0001 for the Institute for Collaborative Biotechnologies. This work was also supported by the National Science Foundation award \#2048122. The views and conclusions contained in this document are those of the authors and should not be interpreted as representing the official policies, either expressed or implied, of the U.S. Government. The U.S. Government is authorized to reproduce and distribute reprints for Government purposes notwithstanding any copyright notation herein.

\bibliography{anthology,custom}
\bibliographystyle{acl_natbib}

\appendix

\newpage

\section{Appendix}

\subsection{Experiment Details}

\paragraph{Pretraining}
We pre-train the mapping network for GPT-2-base~\citep{Radford2019LanguageMA} on the MSCOCO~\citep{Lin2014MicrosoftCC} dataset with 414,113 (image, text) pairs for training.
We pre-train the mapping network for BART-base~\citep{lewis-etal-2020-bart} on VIST~\citep{huang-etal-2016-visual} story-in-sequence subset, with 141,593 (image, text) pairs for training after excluding the images that the users have removed.
For each setting, we pre-train the mapping network for 5 epochs with a batch size of 128, learning rate of 2e-5, weight decay of 0.01, and warmup steps of 5,000.

\paragraph{Few-Shot Training for Downstream Tasks}
Table~\ref{tab:appendix_hyperparameter} lists out the hyperparameters we used during few-show experiments on the three open-ended text generation tasks.

\begin{table}[!htp]\centering
\begin{adjustbox}{width=\linewidth,center}
\begin{tabular}{lccc}\toprule
\textbf{Hyperparameters} &\makecell[l]{\textbf{Concept-to-Text}} &\makecell[c]{\textbf{Text}\\\textbf{Completion}} &\makecell[c]{\textbf{Story}\\\textbf{Generation}} \\\cmidrule{1-4}
Base LM &BART-base &GPT2-base &GPT2-base \\
Batch Size &8 &8 &8 \\
Training Epoch &20 &20 &20 \\
$N_{\text{no\_contra}}$ &4 &10 &15 \\
$\lambda$ &1.5 &1 &0.2 \\
Learning Rate &2e-5 &2e-5 &2e-5 \\
Weight Decay &0.01 &0.01 &0.01 \\
Warmup Steps &400 &400 &400 \\
Max Output Length &64 &100 &150 \\
Num of Beam &10 &10 &10 \\
\bottomrule
\end{tabular}
\end{adjustbox}
\caption{Hyperparameter settings for few-shot open-ended text generation.}
\label{tab:appendix_hyperparameter}
\end{table}

\paragraph{Parameter Search}
We tried the learning rate in the following setting: \{1e-5, 2e-5, 5e-5, 1e-4\}, and tried the batch size in \{4, 8, 16, 32\}.

\paragraph{Parameter Size}
Table~\ref{tab:appendix_parameter_size} lists out the parameter size for the network modules used in our study.


\begin{table}[!htp]\centering
\begin{adjustbox}{width=\linewidth,center}
\begin{tabular}{llc}\toprule
\textbf{Task} &\textbf{Model} &\textbf{Prameter Size} \\\midrule
\multirow{3}{*}{Sentence Completion} &StoryEndGen &~~11M \\
&GPT-2 base &117M \\
&GPT-2 base$+$\methodname~ &160M \\
\midrule
\multirow{3}{*}{Story Generation} &Action-Plan &~~43M \\
&Plan-and-Write &~~34M \\
&SimCTG &117M \\
\midrule
\multirow{5}{*}{Concept-to-Text} &BART-base &110M \\
&KGBART &439M \\
&ModelAdapt &110M \\
&Imagine-and-Verbalize &880M \\
&BART-base$+$\methodname~ &153M \\
\bottomrule
\end{tabular}
\end{adjustbox}

\caption{Parameter sizes of the network modules used in our study.}
\label{tab:appendix_parameter_size}
\end{table}

\begin{figure*}[t]
    \centering
    \includegraphics[width=0.9\linewidth]{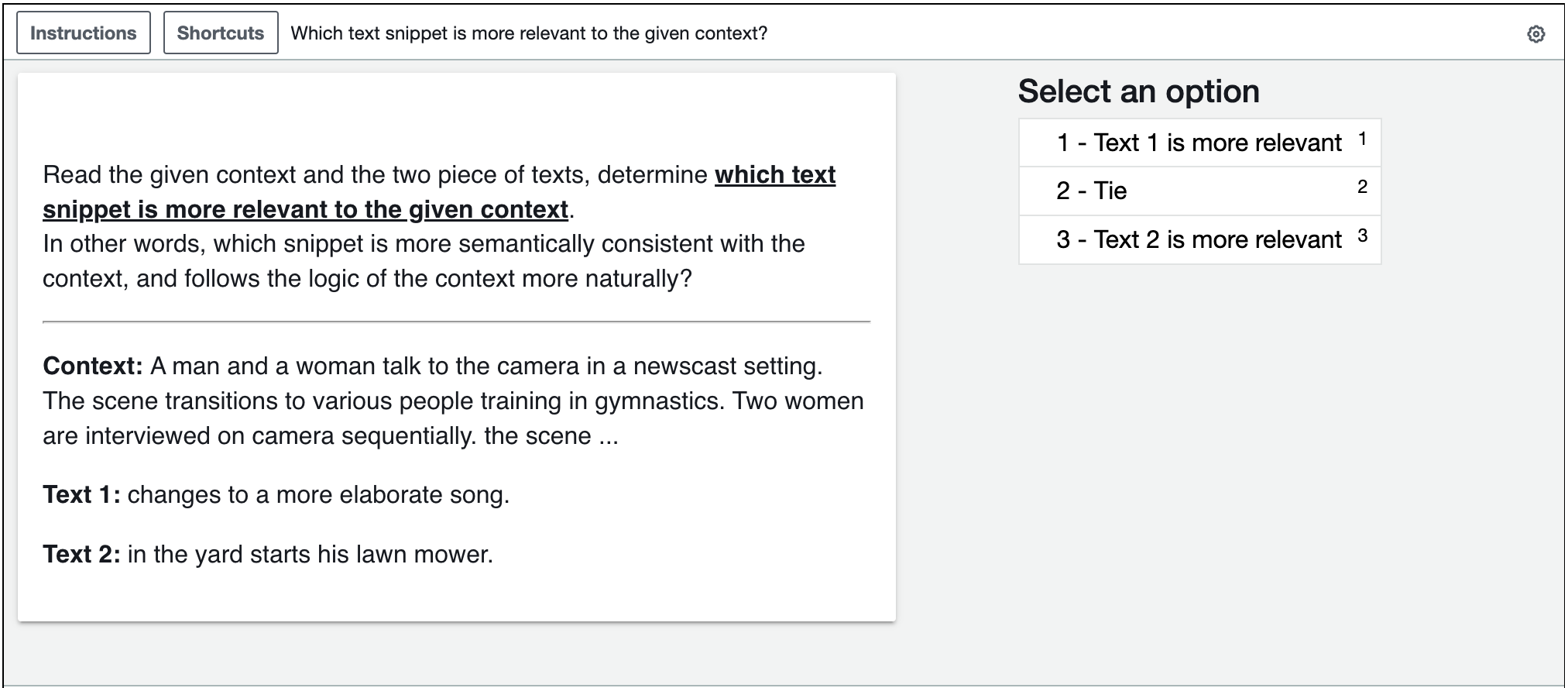}
\caption{
A screenshot of the Amazon Mechanical Turk interface for our human evaluation on text coherency.
}
\label{fig:appendix_human_eval_template}
\end{figure*}

\paragraph{Environment \& Run Time}
Table~\ref{tab:appendix_runtime} lists out the execution time for the three open-ended text generation tasks with 1\% of the training data.
Experiments are conducted on NVIDIA A100.
\begin{table}[!htp]\centering
\begin{adjustbox}{width=0.7\linewidth,center}
\begin{tabular}{lrrr}\toprule
\textbf{Dataset}&\textbf{Text-only} &\textbf{$+$ iNLG} \\\cmidrule{1-3}
ActivityNet &50min &70min \\
ROCStories &70min &95min \\
CommonGen &40min &55min \\
\bottomrule
\end{tabular}
\end{adjustbox}
\caption{The average execution time for one single run (training + inference) on each dataset. Text generation experiments are conducted on NVIDIA A100.}
\label{tab:appendix_runtime}
\end{table}

\subsection{Human Evaluation}
We invite Amazon Mechanical Turk\footnote{\href{https://www.mturk.com/}{https://www.mturk.com/}} annotators to judge the quality of the generated text. Figure~\ref{fig:appendix_human_eval_template} shows an example template we use for head-to-head comparison.

\subsection{More Showcases}
\label{sec:appdix:showcase}

\begin{figure}[h]
    \centering
    \includegraphics[width=0.9\linewidth]{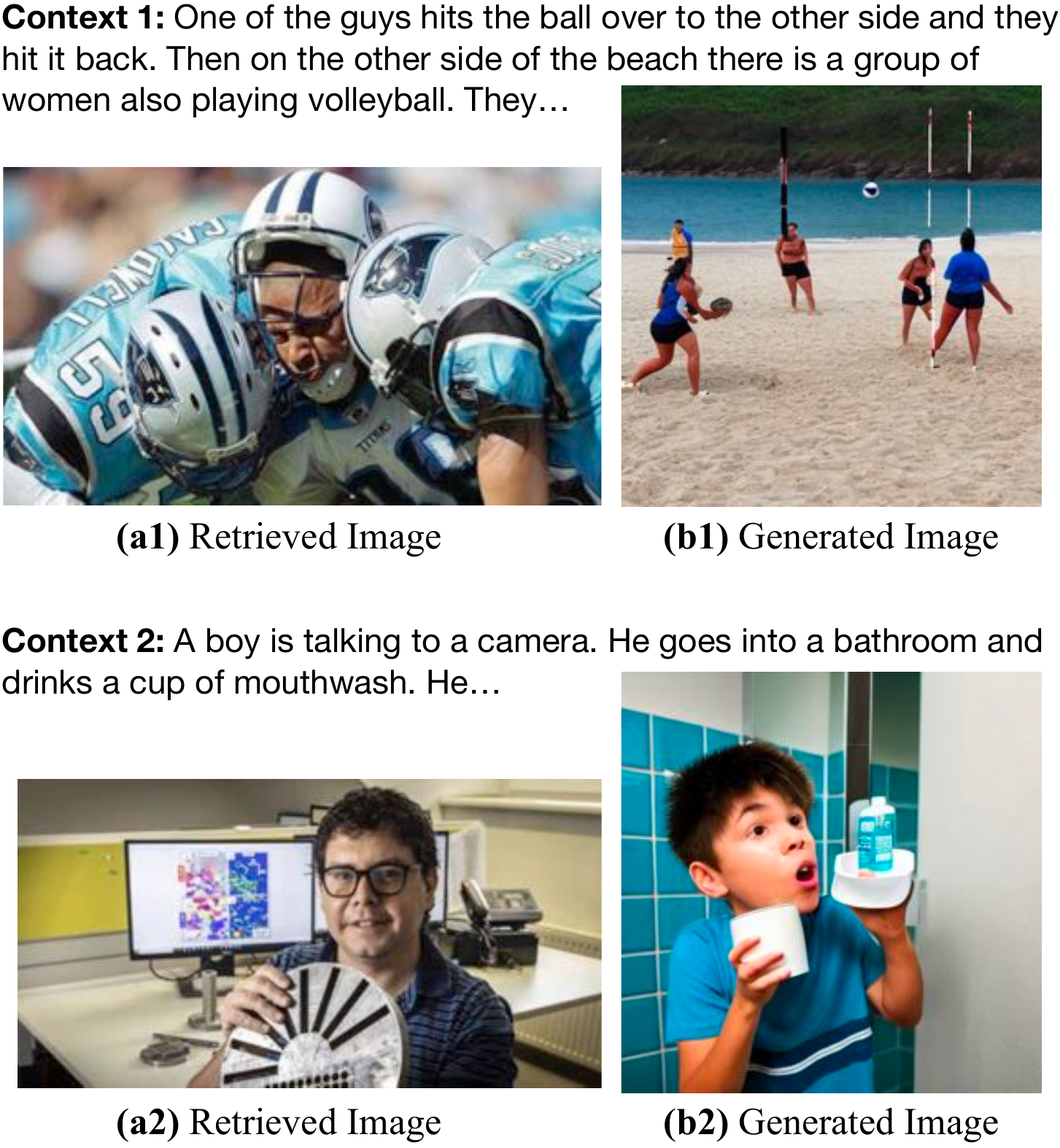}
\caption{
With the context as input, (a1)(a2) is the first returned image by the Yahoo image search engine,\footnotemark and (b1)(b2) is generated by StableDiffusion-\texttt{v1-1}~\citep{Rombach2022HighResolutionIS}. The two input contexts are from the ActivityNet~\citep{Heilbron2015ActivityNetAL} subset in HellaSwag~\citep{zellers-etal-2019-hellaswag}.
}
\label{fig:retrieved_vs_generated}
\end{figure}

\begin{figure*}[ht!]
\centering
	\begin{subfigure}[t]{\textwidth}
		\centering
		\includegraphics[width=0.85\textwidth]{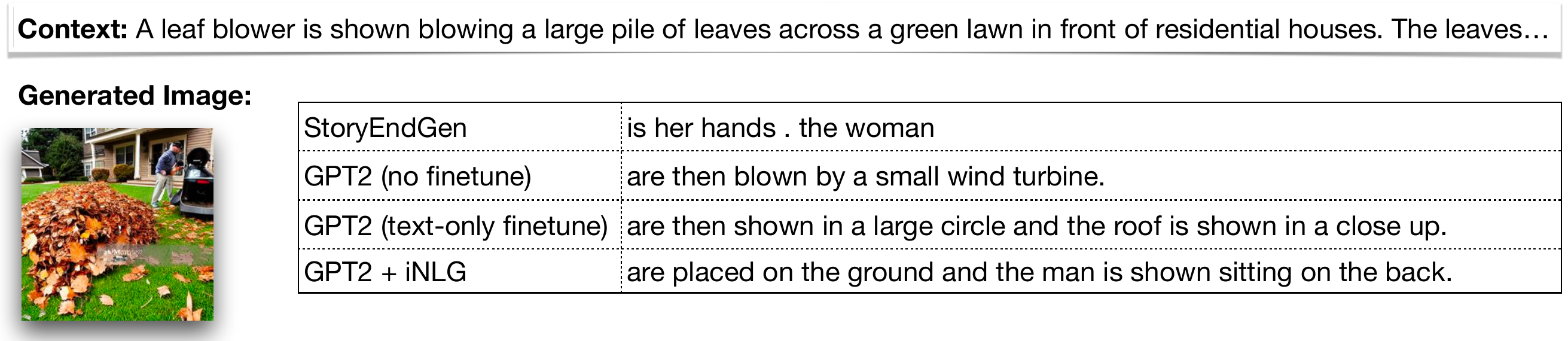}
		\caption{}
	\end{subfigure}%

	\begin{subfigure}[t]{\textwidth}
		\centering
		\includegraphics[width=0.85\textwidth]{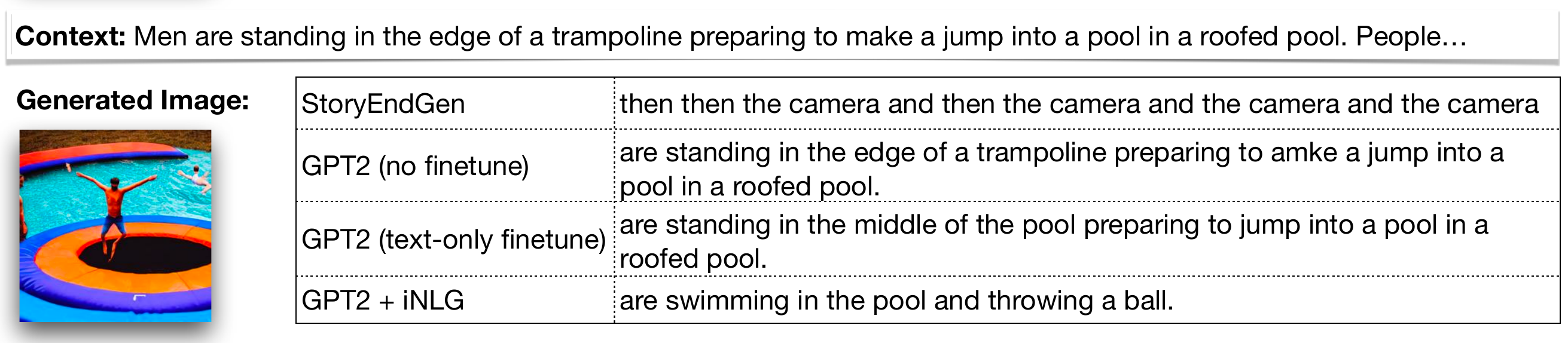}
		\caption{}
	\end{subfigure}%

	\begin{subfigure}[t]{\textwidth}
		\centering
		\includegraphics[width=0.85\textwidth]{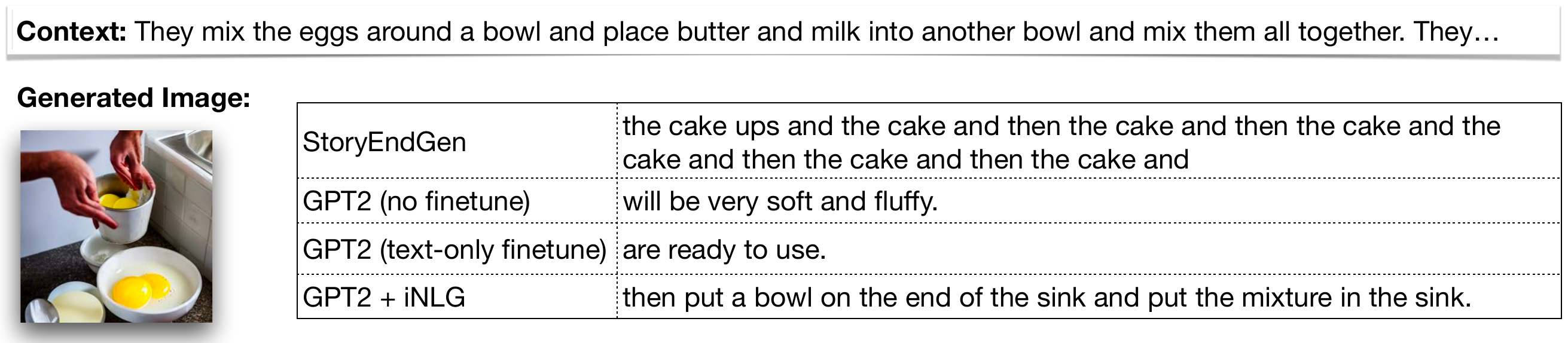}
		\caption{}
	\end{subfigure}%
	
\caption{
Comparisons on few-shot sentence completion performance on ActivityNet.
}
\label{fig:appendix_showcase_activitynet}
\end{figure*}

\begin{figure*}[ht!]
\centering
	\begin{subfigure}[t]{\textwidth}
		\centering
		\includegraphics[width=\textwidth]{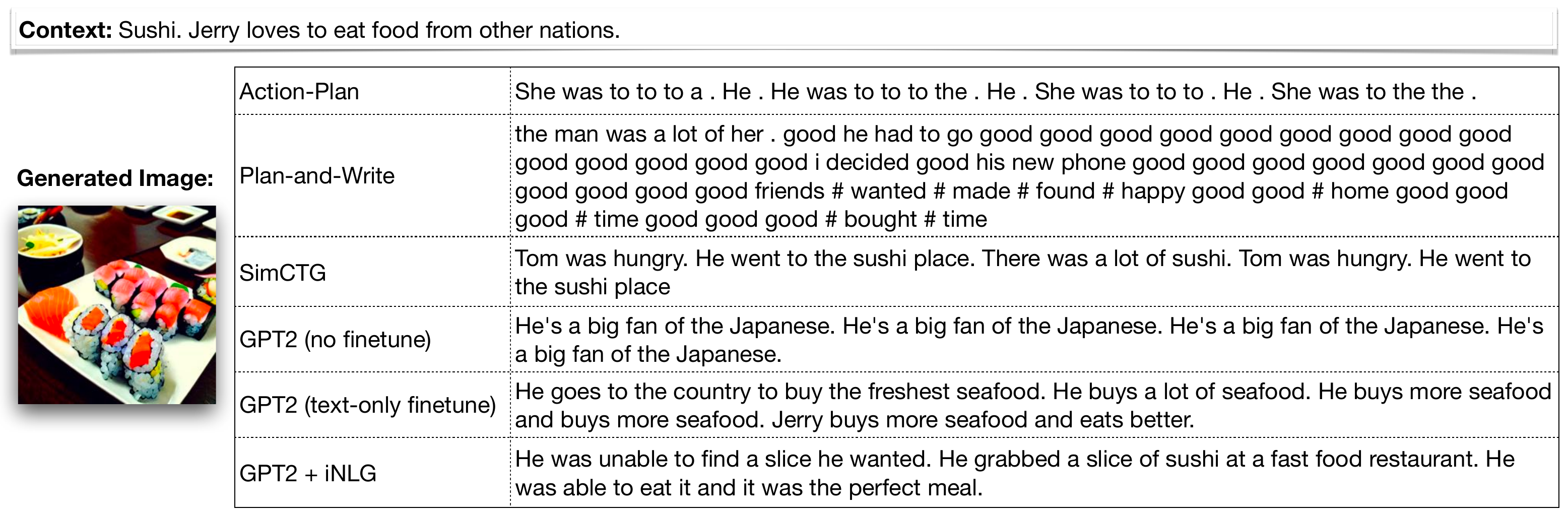}
		\caption{}
	\end{subfigure}%

	\begin{subfigure}[t]{\textwidth}
		\centering
		\includegraphics[width=\textwidth]{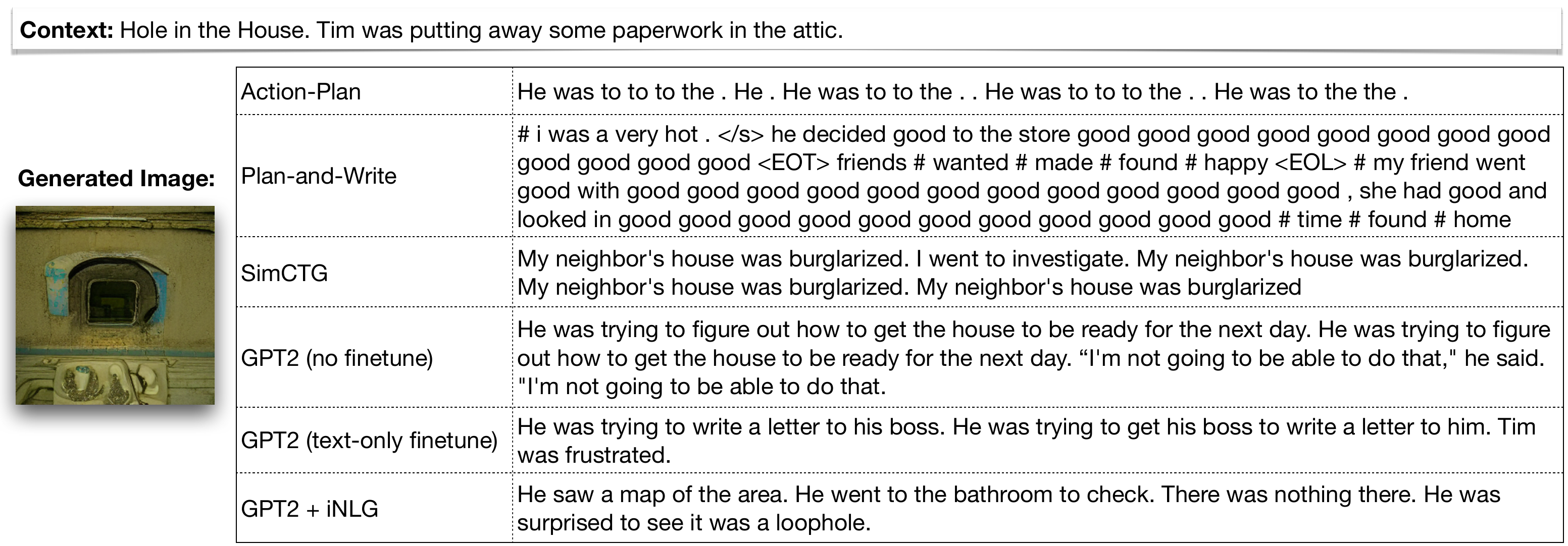}
		\caption{}
	\end{subfigure}%
	
\caption{
Comparisons on few-shot story generation performance on ROCStories.
}
\label{fig:appendix_showcase_rocstories}
\end{figure*}

Figure~\ref{fig:retrieved_vs_generated} compares the images retrieved from Yahoo Image Search and the images generated by StableDiffusion-\texttt{v1-1}~\citep{Rombach2022HighResolutionIS}, which is the text-to-image generation model we used in this work.
Figure~\ref{fig:appendix_showcase_activitynet} and Figure~\ref{fig:appendix_showcase_rocstories} show more examples comparing the sentence endings and stories generated by different models.

\footnotetext{
The screenshots of the search results returned by Yahoo Image Search as of Feb.3rd 2023:
\href{https://github.com/VegB/iNLG/blob/dev/images/yahoo_screenshot_1.png}{link1}, 
\href{https://github.com/VegB/iNLG/blob/dev/images/yahoo_screenshot_2.png}{link2}.
}

\end{document}